\documentclass{article}
\usepackage[T1]{fontenc}
\usepackage[utf8]{inputenc}
\usepackage{float}
\usepackage{calc}
\usepackage{textcomp}
\usepackage{graphicx}
\usepackage{subscript}
\usepackage[unicode=true,
 bookmarks=false,
 breaklinks=false,colorlinks=false]
 {hyperref}

\makeatletter

\usepackage[final]{nips_2017}
\usepackage{amsmath}

\newcommand{\vct}[1]{\boldsymbol{#1}}

\usepackage{url}
\usepackage{amsfonts}
\usepackage{nicefrac}
\usepackage{microtype}
\usepackage{calc}

\title{Human Trajectory Prediction using Spatially aware Deep Attention Models}

\newcommand*\samethanks[1][\value{footnote}]{\footnotemark[#1]}
\author{
  Daksh Varshneya \\
  IIIT-B\thanks{International Institute of Information Technology, Bangalore, KA, 560100}\\
  \texttt{daksh.varshneya@iiitb.org} \\
   \And
   G. Srinivasaraghavan \\
  IIIT-B \samethanks\\
  \texttt{gsr@iiitb.ac.in} \\
}

\@ifundefined{showcaptionsetup}{}{%
 \PassOptionsToPackage{caption=false}{subfig}}
\usepackage{subfig}
\makeatother

\begin{document}

\maketitle
\begin{abstract}
Trajectory Prediction of dynamic objects is a widely studied topic
in the field of artificial intelligence. Thanks to a large number of
applications like predicting abnormal events, navigation system for
the blind, etc. there have been many approaches to attempt learning patterns of motion directly from data using a wide variety
of techniques ranging from hand-crafted features to sophisticated
deep learning models for unsupervised feature learning. All these
approaches have been limited by problems like inefficient features
in the case of hand crafted features, large error propagation across
the predicted trajectory and no information of static artefacts around
the dynamic moving objects. We propose an end to end deep learning model to learn the motion patterns of humans
using different navigational modes directly from data using the much popular sequence to sequence model coupled with a
soft attention mechanism. We also propose a novel
approach to model the static artefacts in a scene and using these to predict the dynamic trajectories. The proposed method, tested on trajectories of pedestrians, consistently outperforms previously
proposed state of the art approaches on a variety of large scale data sets. We also show how our architecture can be
naturally extended to handle multiple modes of movement (say pedestrians, skaters, bikers and buses) simultaneously.
\end{abstract}

\section{Introduction}

Learning and inference from visual data have gained tremendous prominence in  Artificial Intelligence research in recent
times. Much of this has been due to the breakthrough advances in AI and Deep Learning that have enabled vision and image
processing systems to achieve near human precision on many complex visual recognition tasks.
In this paper we present a method to learn and predict the dynamic spatio-temporal behaviour of people moving using
multiple navigational modes in crowded scenes. As humans we possess the ability to effortlessly navigate ourselves in
crowded areas while walking or driving a vehicle. Here we propose an end-to-end Deep Learning
system that can learn such constrained navigational behaviour by considering multiple influencing factors such as the
neighbouring dynamic subjects and also the spatial context in which the subject is. We also show how our
architecture can be naturally extended to handle multiple modes (say pedestrians, skaters, bikers and buses)
simultaneously.

\section{Problem Statement}

We formulate the problem as a learning-cum-inferencing task. The question we seek to answer is ``\emph{having observed
the trajectories $[\vct{x}_i^{(1)},\ldots,\vct{x}_i^{(T)}], 1\leq i\leq N$ of $N$ moving subjects for $T$ time units,
where are each of these subjects likely to be at times $T_1,\ldots,T_n$ after the initial observation time $T$?}'' where
$\vct{x}_i^{(t)}$ denotes the 2-dimensional spatial co-ordinate vector of the $i^{th}$ target subject at time $t$. In
other words, we need to predict values of $[\vct{x}_i^{(T_1)},\ldots,\vct{x}_i^{(T_n)}]$. We assume we have annotations of the
spatial tracks of each unique subject that is participating in the scene.

\section{Related Work}

\subsection{RNNs for Sequence to Sequence modeling}

\noindent Encoder Decoder Models \cite{key-1,key-2,key-3,key-4} were initially introduced for machine translation tasks in \cite{key-5}
followed by automated question answering in \cite{key-16}. These architectures
encode the incoming sequential data into a fixed size hidden representation
using a Recurrent Neural Network(RNN) and then decode this hidden representation using another
RNN to produce a sequentially temporal output.

\noindent These networks have also been modified to introduce
an attention mechanism into them. These networks are inspired by the
attention mechanism that humans possess visually. We as humans, adjust
our focal point over time to focus more at a specific region of our
sight to a higher resolution and the surrounding area to a lower resolution. Attention Networks have been used very successfully for automatic fine-grained image description/annotation \cite{key-7}.

\subsection{Object-Object Interaction modeling}

\noindent Helbing and Molnar's social force model\cite{key-17} was the first
to learn interaction patterns between different objects
such as attractive and repulsive forces. Since then, several variants such as (i) agent based modeling\cite{key-18} to use human
attributes as model priors to learn behavioral patterns, and (ii) feature engineered
approaches like that of Alahi \textit{et.al.}\cite{key-19} which extract social affinity
features to learn such patterns, have been explored.

\noindent Other approaches include finding the most common object
behaviour by means of clustering\cite{key-14,key-15}. Giannotti \textit{et. al.}\cite{key-14}
analyzed GPS traces of several fleets of buses and extracted patterns
in trajectories as concise descriptions of frequent behaviour, both temporally and spatially.

\noindent Most recently, Alahi \textit{et.al.}\cite{key-22} captured the interactions between pedestrians using multiple
Long Short Term Memory Networks(LSTMs)\cite{key-23} and a social pooling mechanism to capture human-human interactions.
While they captured this dynamic interaction, their model failed to understand the static spatial semantics of the
scene. Such spatial modeling exists in \cite{key-20,key-21} but these do not include the dynamic modeling of the crowd.

None of the above approaches naturally extend to multiple classes of moving subjects.

\subsection{Spatial Context modeling}

Earlier works include matching-based approaches\cite{key-8,key-9} which rely on keypoint feature matching techniques. These are slow to compute since they need to match each test image with a database of images. Also, since it is a direct matching approach there is no semantic understanding of the scene. Most of the conventional approaches tend to be brittle since they rely heavily on hand-crafted features.

\noindent A deep learning approach was also used for spatial context modeling in \cite{key-10}. This work hypothesizes
that dynamic objects and static objects can be matched semantically based on the interaction they have between each
other.
The authors assume that a random image patch of the scene contains enough evidence, based on which the discrimination
between likely patches and unlikely patches can be made for a particular object. They did not explore the possibility of
adding additional static context around the patches to augment the model. We build on this distinction later on in
section 4.1.

\section{Proposed System}
\noindent This section describes our solution in detail and is organized as follows. Firstly,
we describe our proposed \textbf{Spatially Static Context Network}(SSCN)
to model the static spatial context around the subject of interest. Next, we define the \textbf{Pooling
Mechanism} which captures the influence, the neighboring
subjects and the nearby static artefacts have on a target subject.
Next, we describe our complete model which uses an attention mechanism
with LSTMs to learn patterns from spatial co-ordinates
of subjects while preserving the spatial and dynamic context around
the subjects of interest. Though LSTMs have traditionally been used to model (typically short-term) temporal
dependencies in sequential data, their use along with an appropriate attention mechanism enables us to use them for
tasks like trajectory planning that requires long-term dependency modeling. We also show how this extends in
principle to multiple classes of moving subjects.

\subsection{Spatial Context Matching}
\label{sec:sscn}
\noindent Modeling the spatial context in a given scene is a challenging task since it should be semantically  representative and also highly discriminative. It is also very important for the model to generalize to a variety of complex scenes and enable inferences about human-space interactions in them.

\noindent Our proposed architecture is composed of Convolutional Neural Networks(CNNs)\cite{key-24} and is inspired by
the Spatial Matching Network introduced in \cite{key-10}. Although, our architecture is very similar to the one proposed
in this work, we differ significantly from this approach in two ways. First, it is redundant to build an input branch
which takes image patches of different objects because an object belonging to the same semantic class (car, pedestrian,
bicyclist) should ideally have the same spatial matching score with any particular random patch of the image. For
example, any car should have the same matching score with a patch of road and hence we do not need to differentiate
between different cars.

\noindent Second, it might be difficult for a network to look at
a small scene patch and infer the matching score from it. For instance,
a trained CNN might have learned different textures of different static
artefacts like that of a road or a pavement, but such textures could
occur anywhere in the image. For example, the texture of the roof
in a particular scene could match the texture of the pavement in a different scene image. So it is important for
the network to have information about the larger context or the region
surrounding the input patch. 
This will help the network to generalize better across different scenes and have a better semantic understanding of different complex scenes.

\noindent We incorporate our hypothesis in our proposed network shown
in figure 1. We call this network as the \textbf{Spatially Static
Context Network(SSCN)}. The network has three input streams - \textbf{S}ubject
stream, \textbf{P}atch stream and \textbf{C}ontext stream. 

\begin{figure}[h]
\centering %
\fbox{\includegraphics[scale=0.35]{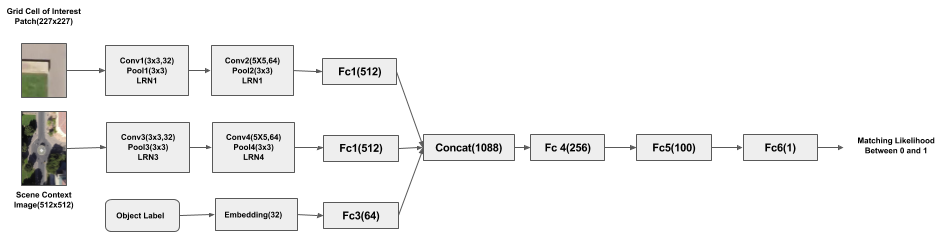}} \caption{Spatially Static Context Network Architecture}
\end{figure}

\noindent For the \textbf{s}ubject input stream, we input a class label $s\in
\{0,1,2,\ldots\}$ to indicate whichever semantic class the dynamic subject
belongs to. This input is passed through an embedding layer followed
by a dense fully connected layer.

\noindent For the \textbf{p}atch stream the input image is that of the grid
cell of interest $\rho$. To incorporate a local context around
that patch, we also take the part of the image surrounding that cell. Hence,
we take the size of the grid cell of interest as $g_{\rho}\times g_{\rho}$ and add an annulus of width $g_{\rho}$ around
it (we appropriately pad the original image in order to add such a context for the cells on the boundary of the image).
Thus, the final size of input patch becomes $3g_{\rho}\times 3g_{\rho}$. This is later re-sized to $d_P\times d_P$ for a
fixed $d_P$. We pass this input patch through a stack of convolutional layers along with pooling and local response normalization after each
layer followed by a dense fully connected layer as shown in figure
1.

\noindent The third stream, that captures the overall image \textbf{c}ontext, is in line with our
second hypothesis. This stream of the network is incorporated to introduce
a global context around the image patch of interest. The input
to this stream $\zeta$ is the whole scene image re-sized to a fixed
dimension size of $d_I\times d_I$. We again use another stack of convolutional
layers along with pooling and local response normalization after each
layer followed by a dense fully connected layer to extract hierarchical
features of the scene image.

\noindent To train such a network, we merge all the input streams by concatenating the outputs of their respective dense
fully connected layers and add a stack of 2 fully connected layers followed by a fully-connected output layer. The
output layer consists of just a single sigmoid neuron, giving the likelihood of a subject of type $s$  stepping on the
given patch $\rho$.
The ground truth likelihood we use for training is the value associated with the grid cell of interest (not the complete
input patch).

\noindent We train the model over all triplets $(s,\rho,\zeta)\in\mathcal{D}$ of subject type, patch and image, by
minimizing the cross entropy between the actual likelihood $\mu_{s\rho\zeta}$ and the
predicted likelihood $\hat{\mu}_{s\rho\zeta}$ - 
\[
H_{\mathcal{D}} = -\frac{1}{|\mathcal{D}|}\sum_{(s,\rho,\zeta)\in
\mathcal{D}}\left(\mu_{s\rho\zeta}log(\hat{\mu}_{s\rho\zeta})+(1-\mu_{s\rho\zeta})log(1-\hat{\mu}_{s\rho\zeta})\right)
\]
The ground truth likelihood value for each patch is computed by counting the frequency of unique subjects of type $s$
occupying the patch at some time during the course of their individual trajectories and dividing it by the total number of unique
subjects of the same type in the scene. Note that $\mathcal{D}$ is constructed from the annotated videos by tracking and
counting the incidence of each subject in the video with every patch in a subsample of video frames.

\subsection{Pooling Mechanism}
Our model was trained on a subsample (one in every ten frames) of frames from the videos that formed our training
dataset. We describe how we pool the static and dynamic contexts for a given frame $\mathcal{F}$ from the sample along
with the representation of the historical trajectory. We use the subscript $\mathcal{F}$ to indicate the fact that the
pooling being done is specific to the frame.

\subsubsection*{Dynamic Context Pooling}

\noindent Humans moving in a crowded area adapt their motion based
on the behaviour of the people around them. For example, pedestrians often
completely alter their paths when they see someone else or a group of people approaching them. Such behaviour cannot be predicted by observing
a pedestrian in isolation without considering the surrounding dynamic and static context. This behaviour motivated the pooling mechanism of the Social
LSTM model\cite{key-22}.  We borrow the same pooling mechanism to capture such influences from neighbouring subjects for our model.

\noindent We use LSTMs to learn an efficient hidden representation of the temporal behaviour of subjects as part of
the encoder.
Since these hidden representations would capture each subject's behaviour until the observed time step, we can use
these representations to capture the influence that the neighbouring subjects would have on a target subject.

\noindent We consider a spatial neighbourhood of size $\left(d_s\times d_s\right)$ around each moving subject (again
with appropriate padding for the boundary cells) which in turn is subdivided into a $(g_s\times g_s)$ grid with each
grid cell of size $\left(\frac{d_s}{g_s}\times\frac{d_s}{g_s}\right)$.
Let $\vct{h}_{js\mathcal{F}}^{(t)}$ denote the LSTM encoded hidden representation (a vector of dimension $d_H$) of $j^{th}$ subject
of type $s$ at time $t$ (when the current frame is $\mathcal{F}$). Also let $C$ be the number of subject classes.
We construct \emph{social tensors}\cite{key-22} $\mathcal{S}_{i\mathcal{F}}^{(t)}$ each of size $(g_s\times g_s\times
d_H\times C)$ that capture the social context in a structured way

\noindent 
\[
\mathcal{S}_{i\mathcal{F}}^{(t)}(r,c,:,:) = \left[
\sum_{j=1}^{N_s}\mathbb{I}_{irc\mathcal{F}}^{(t)}[s,j].\vct{h}_{js\mathcal{F}}^{(t-1)} \right]_{s=1}^C
\]
where $N_s$ is the total number of subjects of type $s$ and $\mathbb{I}_{irc\mathcal{F}}^{(t)}[s,j]$
is an indicator function which denotes whether the $j^{th}$ subject of type $s$
is in the $(r,c)^{th}$ grid cell of the spatial neighbourhood of the $i^{th}$ subject at time $t$.

\subsubsection*{Static Context Pooling}

As described earlier, the SSCN model is designed to predict the likelihood of
a subject like a pedestrian stepping on a specific input image patch,
given the larger context around the patch and the scene
itself. We use this to provide a surrounding context for each subject from its current position which in turn
influences the next position of the subject.

We first build a spatial map for each subject class and location in a frame of the video, with the probabilities of a
subject of that class ever visiting that location. This map is built offline using the pretrained SSCN network described in Section
\ref{sec:sscn}. Given a subsample of frames, for every frame $\mathcal{F}$ in the sample, patch $\rho$ and subject class
$s$ we build the map 
$\mathcal{M}_{s\rho}^{\mathcal{F}} = SSCN(s,\rho, \mathcal{F})$. 
We extract a \emph{Reachability Tensor} $R^{(t)}_{i\mathcal{F}}$ for the $i^{th}$ subject (say of class $s$) at time $t$
from the static context map $\mathcal{M}_{s\rho}^{\mathcal{F}}$. Given its current position $\vct{x}_i^{(t)}$, let $\Phi_i^{(t)}$
be the collection of all patches of size $(3g_p\times 3g_p)$ that are centered at $(g_p\times g_p)$ patches at most $\left(\frac{d_R}{2}\right)$ away
on each axis from $\vct{x}_i^{(t)}$. We basically construct a reachability context tensor accounting for a patch of size
$(d_R\times d_R)$ with $\vct{x}_i^{(t)}$ at the center. The extracted reachability context tensor is therefore
\[ R^{(t)}_{i\mathcal{F}} = \left[SSCN(s,\rho, \mathcal{F})\right]_{\rho\in\Phi_i^{(t)}} \]

\subsection{Spatio - Temporal Attention Model}

\noindent In our work, we take the encoder-decoder architecture as the base model and apply a soft attention mechanism on top of it. The motivation for applying an attention mechanism
is straight-forward. Subjects often change their pre-panned trajectories suddenly when the 'context' changes.
Imagine a pedestrian in an airport walking towards the
security, suddenly realizing that he/she needs to pick up the baggage tag and as a result making a sharp
course-correction to move towards the check-in counter.
Since, the model proposed by Alahi \textit{et. al.}\cite{key-22} only takes the last
time step hidden representation of the pedestrian of interest, the
model will be responsive to immediate instincts like collision avoidance
but not be very useful in long term path planning. Moreover, once
the model starts its predictions, even a small error in the prediction
could mean that the erroneous hidden representations are propagated to future time steps.

\noindent The full architecture for our spatio-temporal attention
model is shown in figure 2.

\begin{figure}[h]
\centering %
\fbox{\includegraphics[scale=0.28]{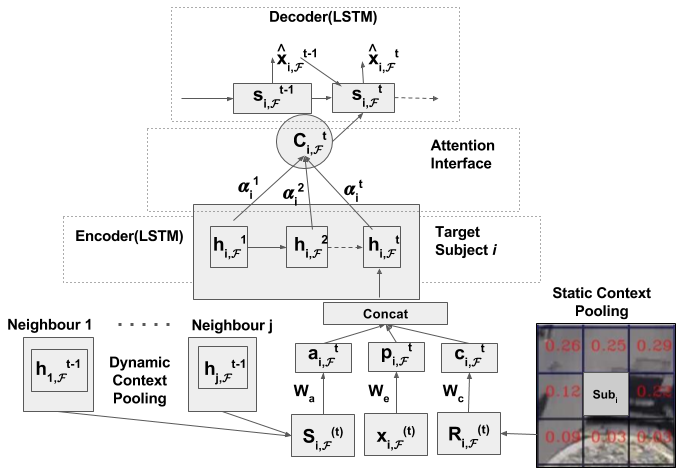}} \caption{Spatio-Temporal Attention Model Architecture}
\end{figure}

We first embed the spatial coordinate
input $\vct{x}_{i\mathcal{F}}^{(t)}$, the dynamic context pooled tensor $S_{i\mathcal{F}}^{(t)}$ and static
reachability tensor $R_{i\mathcal{F}}^{(t)}$ to fixed dimensions using three separate sigmoid embedding layers.
\[
p_{i\mathcal{F}}^{t}=\phi(\vct{x}_{i\mathcal{F}}^{(t)},W_{e}),
~~~a_{i\mathcal{F}}^{t}=\phi(S_{i\mathcal{F}}^{t},W_{a}),~~~c_{i\mathcal{F}}^{t}=\phi(R_{i\mathcal{F}}^{t},W_{c})
\]
where $W_{e},W_{a}$ and $W_{c}$ are the embedding matrices and $\phi$ is
the sigmoid function. Next, the three embeddings $p_{i\mathcal{F}}^{t}$, $a_{i\mathcal{F}}^{t}$
and $c_{i\mathcal{F}}^{t}$ are concatenated to form the input to the encoder.
The encoder outputs a fixed size hidden state representation $\vct{h}_{i\mathcal{F}}^{t}$
at each time step t, 
\[
\vct{h}_{i\mathcal{F}}^{t}=LSTM(\vct{h}_{i\mathcal{F}}^{t-1},p_{i\mathcal{F}}^{t},a_{i\mathcal{F}}^{t},c_{i\mathcal{F}}^{t},W_{enc})
\]
where $\vct{h}_{i\mathcal{F}}^{t-1}$ is the hidden state representation output of the
decoder at the last timestep $(t-1)$ and LSTM is the encoding function
for the encoder with weights $W_{enc}$.

\noindent We use a context vector $C_{i\mathcal{F}}^{(t)}=\sum_{j=t-k+1}^t\alpha^{(j)}_{i}.\vct{h}^{(j)}_{i\mathcal{F}}$
that depends on the encoder hidden states occuring in a fixed size temporal attention
window of size $k$ ---
$[\vct{h}_{i\mathcal{F}}^{(t-k+1)},\ldots.,\vct{h}_{i\mathcal{F}}^{(t-1)},\vct{h}_{i\mathcal{F}}^{(t)}]$.
This makes it possible for the network to dynamically generate different
context vectors at different timesteps with the knowledge of $k$ encoded
states back in time. We follow the previously proposed
attention mechanism like in Bahdanau \textit{et. al.}\cite{key-6} to compute the attention weights $\alpha^{(j)}_{i}$.
This context vector
is generated by the attention mechanism which puts an emphasis over
the encoder states and generates a 'behaviour context' for the subject
of interest.

\noindent The context vector $C_{i\mathcal{F}}^{(t)}$ feeds into the decoder unit
to generate the decoder states. The decoder in turn generates a 5-tuple as its output representing the parameters of
a bivariate Gaussian distribution over the predicted position of the subject at the next time step. Denoting the decoder
state at time $t$ (for the frame $\mathcal{F}$) as $\vct{s}^{(t)}_{i\mathcal{F}}$, the decoder state and the
predicted bivariate Gaussian model
$\vct{\theta}_{i\mathcal{F}}^{(t)}\equiv\mathcal{N}(\vct{\mu}_i^{(t)},\vct{\sigma}_i^{(t)},\rho_i^{(t)})$
for the position of the subject $\vct{x}_{i\mathcal{F}}^{(t)}$ at time $t$ are computed as
\[
\vct{s}^{(t)}_{i\mathcal{F}} =
LSTM(\vct{s}^{(t-1)}_{i\mathcal{F}},\vct{x}_{i\mathcal{F}}^{(t-1)}, C_{i\mathcal{F}}^{(t)},W_{dec}),~~~~
\vct{\theta}_{i\mathcal{F}}^{(t)}=ReLU(\vct{s}^{(t)}_{i\mathcal{F}},W_{o})
\]
Note that $\vct{x}_{i\mathcal{F}}^{(t-1)}$, the position at time $(t-1)$, is taken as 
the actual ground truth value at time $(t-1)$ during training and the model predicted value during
inference. The above model can be trained for multiple classes of subjects with a separate model for each individual class.

Our model therefore accounts the current spatial coordinates, social context that incorporates all the moving subjects in the scene and the static context that accounts for the reachability of the subjects across the patches in
the scene.

During inference, the final predicted position $\hat{\vct{x}}_{(t)}^{i}$ is sampled from the predicted distribution
$\vct{\theta}_i^{(t)}$.

\subsection*{Cost Formulation}

\noindent We train the network by maximizing the likelihood of the ground truth position being generated
from the predicted distribution. Hence, we jointly learn all the parameters by minimizing the negative log-Likelihood
loss $L_{i}=-\sum_{t=T_1}^{T_n}\log(P(\vct{x}_i^{(t)}\mid\vct{\mu}_i^{(t)},\vct{\sigma}_i^{(t)},\rho_i^{(t)})$
for the $i^{th}$ trajectory.
An important aspect of the training phase is that, since the LSTM
layers of the encoder and decoder units are shared between all the subjects of a particular type, all parameters of the models have to be learned jointly.
Thus, we back-propagate the loss for each trajectory $i$ of each subject type at every time step $t$.

\section{Experiments}

\subsection*{Dataset}
We use three large scale multi-object tracking
datasets - ETH \cite{key-11} , UCY \cite{key-12} and the Stanford
Drone Dataset \cite{key-13}. The ETH and UCY datasets consist of 5 scenes
with 1536 unique pedestrians entering and exiting
the scenes. It includes challenging scenarios like groups of people walking together, 2 different groups of people crossing each other and also behaviour such as a pedestrian
deviating completely from it's followed path almost instantaneously.

\noindent On the other hand, the Stanford Drone Dataset\cite{key-13} consists of multiple aerial imagery comprising of 8 different
locations around the Stanford campus and
objects belonging to 6 different classes moving around. We use only the
trajectories of pedestrians to train and test our models.

\subsection*{Setup}

We set the hyperparameters of our model using a cross validation strategy following a leave one out approach. For the
SSCN model we take the size of the grid $g_{\rho}$ in the patch stream as 60 and the re-sized input dimension as 227 x 227. The input dimension of
the context stream is set to 512 x 512. The overall network is trained using a learning rate of 0.002, with Gradient
Descent Optimizer and a batch size of 32.

\noindent To compare our results with that of previous state of the art model - S-LSTM\cite{key-22}, we limit the subject type to only pedestrians. We set the common
hyperparameters of the Spatio-Temporal Attention Model to be the same as that of theirs. Trajectories are downsampled so
as to retain one in every ten frames. We observe the trajectories for a period of 8 time steps ($T$) with an
attentional window length k of 5 time steps and predict for future 12 time steps ($n$). We set the reachability distance $d_{R}$ to 60. We also limit the number
of pedestrians in each frame to 40. The model is trained using a learning rate of 0.003 with RMSProp as the optimizer.

\subsubsection*{Evaluation Metrics}

We use two evaluation metrics as proposed in Alahi \textit{et.al.}\cite{key-22} - 
\textbf{(i) Average Displacement error} - The euclidean distance between the
predicted trajectory and the actual trajectory averaged over all time-steps
for all pedestrians and 
\textbf{(ii) Final Displacement error} - The average euclidean distance between
the predicted trajectory point and the actual trajectory point at
the end of $n$ time steps.

\begin{table}[H]
  \caption{Comparison across datasets}
  \label{quantitative result}
  \centering
  \begin{tabular}{llllll}
        \hline
        Metric & Dataset & O-LSTM\cite{key-22} & S-LSTM\cite{key-22} & D-ATT & SD-ATT \tabularnewline
        \hline 
        Avg. Disp. Error & ETH\cite{key-11} & 0.49 & 0.50 & \textbf{0.47} & -\tabularnewline
         & HOTEL\cite{key-11} & \textbf{0.09} & 0.11 & 0.12 & -\tabularnewline
         & ZARA1\cite{key-12} & 0.22 & 0.22 & \textbf{0.18} & -\tabularnewline
         & GATES1\cite{key-13} & 0.16 & 0.12 & 0.11 & \textbf{0.09}\tabularnewline
         & GATES2\cite{key-13} & 0.15 & 0.17 & 0.14 & \textbf{0.10}\tabularnewline
         & GATES3\cite{key-13} & 0.18 & 0.16 & 0.13 & \textbf{0.13} \tabularnewline
        \hline 
        Final. Disp. Error & ETH\cite{key-11} & 1.06 & 1.07 & \textbf{0.85} & -\tabularnewline
         & HOTEL\cite{key-11} & 0.20 & 0.23 & \textbf{0.19} & -\tabularnewline
         & ZARA1\cite{key-12} & \textbf{0.46} & 0.46 & 0.48 & -\tabularnewline
         & GATES1\cite{key-13} & 0.28 & 0.25 & 0.19 & \textbf{0.17}\tabularnewline
         & GATES2\cite{key-13} & 0.40 & 0.37 & 0.38 & \textbf{0.35}\tabularnewline
         & GATES3\cite{key-13} & 0.26 & 0.26 & 0.25 & \textbf{0.24}\tabularnewline
        \hline 
\end{tabular}
\end{table}

\subsection{Quantitative Results}

We build two separate models for the complete problem statement -
one with only dynamic context pooling coupled with the attention mechanism
which we denote as - \textbf{D-ATT} Model and the second with static
context pooling added to the D-ATT model which we denote as - \textbf{SD-ATT}.
We cannot test the SD-ATT model on ETH and UCY datasets as the resolution of the videos in these datasets is too low which makes it impossible to use the SSCN model for static context pooling. Still, we consistently outperform the S-LSTM\cite{key-22} and O-LSTM\cite{key-22} models on both the evaluation
metrics on all three datasets as shown in table 1. We also show that
the results of the SD-ATT model on the Stanford Drone Dataset are
much better than the Social LSTM\cite{key-22} model.

\subsection{Qualitative Results}
We demonstrate scenarios where our models perform better than the S-LSTM\cite{key-22} model. Firstly, in Figure 3 we
show the results for SSCN model. The second column of the figure shows an example of the constructed saliency map with
the shade of blue denoting the likelihood value of the corresponding patch. We can see that the likelihood values are
low near the roundabout, car and trees and high for areas such as road and pavement. In rest of the figures, each unique
pedestrian is depicted by a unique colour. In Figure 4, the first column shows that the SD-ATT model has learned to
predict non linear trajectories as well. The next two columns depict the collision avoidance property learned by the
model. In both the examples, the model either decelerates one of the pedestrian or it diverts it to avoid collision.

\noindent Figure 5(a) compares the predictions of SD-ATT model against the Social LSTM\cite{key-22} model. Since, the
S-LSTM model only considers the last time step's hidden representation, it thinks that the pedestrian wants to take a
turn and hence follows the curved path. On the other hand, our SD-ATT model interprets this as a sharp turn since it has
only seen this change in the behaviour over the last two time steps and hence takes a gradual turn. This demonstrates
the advantage of using an attention mechanism. Figure 5(b) demonstrates the advantage of static spatial pooling. In both
the examples shown, the pedestrian walks straight for $T$ time steps because of which
S-LSTM\cite{key-22} and D-ATT models predict a straight path. On the other hand, SD-ATT model captures a static obstacle
in front of the pedestrian and hence takes a diversion from the followed path.

\begin{figure}[H]
\centering %
\fbox{\includegraphics[scale=0.45]{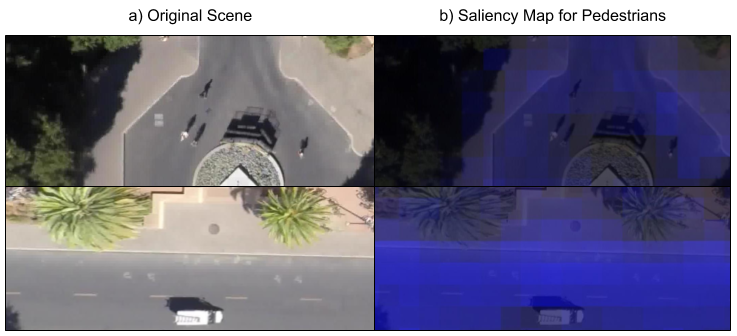}}\caption{Results for SSCN model on Stanford Drone Dataset}
\end{figure}
\vspace*{-3mm}

\begin{figure}[H]
\centering %
\fbox{\includegraphics[scale=0.42]{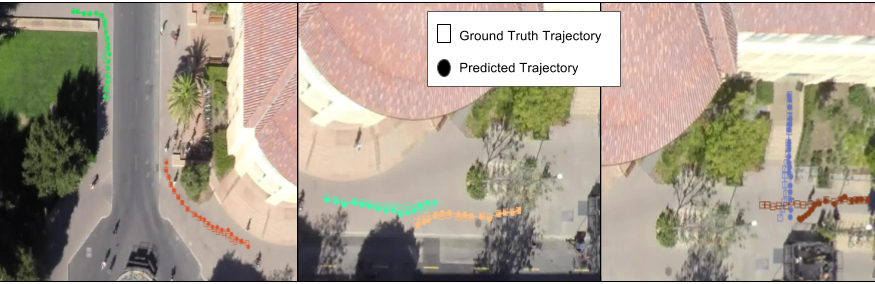}} \caption{Sample predicted trajectories which show the non-linear and collision avoidance property of the model}
\end{figure}
\vspace*{-3mm}
\begin{figure}[h]
\centering
\subfloat[{Advantage of Attention Mechanism}]{\includegraphics[scale=0.18]{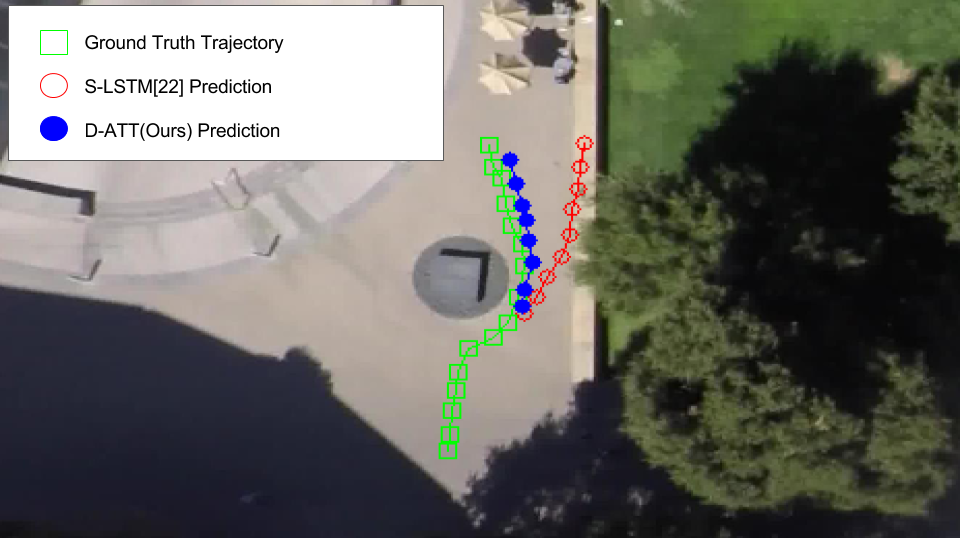}

} 
\vspace*{-3mm}
~\subfloat[Advantage of Static Context Pooling]{\includegraphics[scale=0.22]{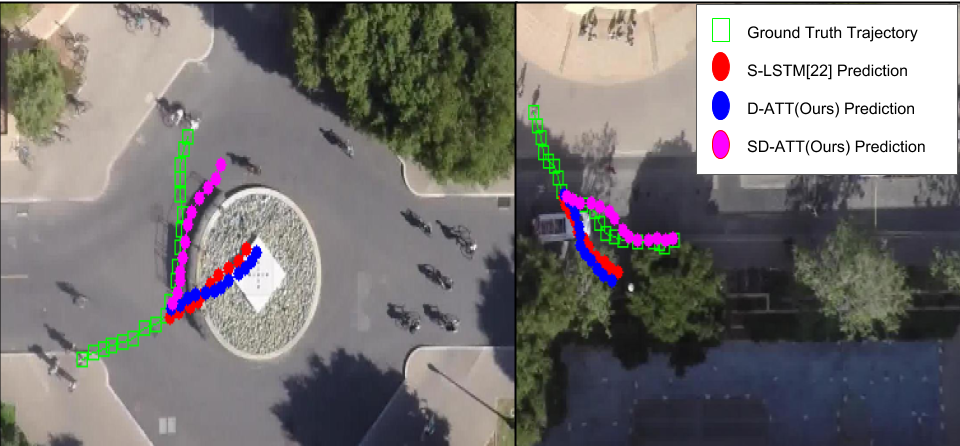}}

\caption{Comparison of our models against S-LSTM\cite{key-22} model}
\end{figure}

\section{Conclusion}
We propose a novel deep learning approach to the problem of human trajectory prediction. Our model successfully extracts motion patterns in an unsupervised manner. Compared to the previous state of the art works, our approach models both dynamic and spatial context around any type of subject of interest which results in better prediction of trajectories. Our proposed method outperforms previous state of the art method on three large scale datasets. In addition to this, we also propose a novel CNN based SSCN architecture which helps in better semantic understanding of the scene. Future work includes evaluating the proposed models for multiple classes of objects.




\bibliographystyle{abbrv}

\end{document}